\documentclass[letterpaper, 10 pt, conference]{ieeeconf}  
\IEEEoverridecommandlockouts                              
\usepackage{times}
\usepackage{multicol}
\usepackage[bookmarks=true]{hyperref}
\usepackage{graphicx} 
\usepackage{amsmath} 
\usepackage{amssymb}  
\usepackage{bm} 
\usepackage{array}
\usepackage{colortbl}	
\usepackage[table]{xcolor}
\usepackage{subfig}

\newcommand{\pder}[2]{\frac{\partial#1}{\partial#2}}		
\DeclareMathOperator*{\lexmin}{\text{lex}\!\min}			
\DeclareMathOperator*{\half}{\frac{1}{2}}					
\newcommand{\mat}[1]{\ensuremath{\begin{bmatrix}#1\end{bmatrix}}}	
\newcommand{\rank}[1]{\text{rank}(#1)}							
\newcommand{\x}{\ensuremath{\times}}
\newcommand{\spac}{\ensuremath{\quad}}						
\newcommand{\T}[0]{\ensuremath{\top}}							
\newcommand{\pinv}[0]{\ensuremath{\dagger}}					
\newcommand{\Rv}[1]{\ensuremath{\mathbb{R}^{#1}}}				
\newcommand{\R}[2]{\ensuremath{\mathbb{R}^{#1\times #2}}}		

\newenvironment{definition}[1][Definition]{\begin{trivlist}
\item[\hskip \labelsep {\bfseries #1}]}{\end{trivlist}}

\title{\LARGE \bf
Partial Force Control of \\Constrained Floating-Base Robots
}

\author{Andrea Del Prete$^{1}$, Nicolas Mansard$^{1}$, Francesco Nori$^{2}$, Giorgio Metta$^{3}$, Lorenzo Natale$^{3}$
\thanks{$^{1}$Del Prete and Mansard are with the LAAS/CNRS, Toulouse, France. {\tt\small adelpret@laas.fr, nmansard@laas.fr}}%
\thanks{$^{2}$Nori is with the RBCS Department, Istituto Italiano di Tecnologia, Genova, Italy. {\tt\small francesco.nori@iit.it}}%
\thanks{$^{3}$Metta and Natale are with the iCub Facility, Istituto Italiano di Tecnologia, Genova, Italy. %
        {\tt\small name.surname@iit.it}}%
}

\begin{document}

\maketitle
\thispagestyle{empty}
\pagestyle{empty}

\begin{abstract}
Legged robots are typically in rigid contact with the environment at multiple locations, which add a degree of complexity to their control.
We present a method to control the motion and a subset of the contact forces of a floating-base robot. 
We derive a new formulation of the lexicographic optimization problem typically arising in multi-task motion/force control frameworks.
The structure of the constraints of the problem (i.e. the dynamics of the robot) allows us to find a sparse analytical solution.
This leads to an equivalent optimization with reduced computational complexity, comparable to inverse-dynamics based approaches.
At the same time, our method preserves the flexibility of optimization based control frameworks.
Simulations were carried out to achieve different multi-contact behaviors on a 23-degree-of-freedom humanoid robot, validating the presented approach.
A comparison with another state-of-the-art control technique with similar computational complexity shows the benefits of our controller, 
which can eliminate force/torque discontinuities.
\end{abstract}

%

\section{Introduction}
Control of floating-base mechanical systems (e.g. legged robots) is still a main concern for the control community.
One of the reasons accounting for this on-going research is that floating-base systems are \emph{underactuated}, hence they cannot be feedback-linearized \cite{Spong1994}.
The problem becomes even more complex when these systems are \emph{constrained}, that is their dynamics is subject to a set of (possibly time-varying) nonlinear constraints. 
This is the typical case for legged robots, whose motion is constrained by rigid contacts with the ground.

Sentis \cite{Sentis2008} and Park \cite{Park2006a} presented a framework for prioritized motion and force control of humanoid robots.
This framework builds on the idea of Operational Space dynamics \cite{Khatib1987}, resulting in a massive use of dynamics quantity such as the joint space mass matrix \cite{featherstone2008rigid}.
Righetti et al. proposed an alternative approach \cite{Righetti2011} based on recent results from analytical dynamics \cite{Aghili2005}.
They projected the robot dynamics into the nullspace of the constraints, using a geometric projector.
The projection cancels the constraint forces from the system dynamics, removing any need of force measurements.
These geometric projectors are faster to compute than those depending on inertial quantities used in \cite{Sentis2008,Park2006a}, 
so the resulting control laws are simpler and computationally more efficient.
Mistry et al. \cite{Mistry2011} presented an in-between approach, extending the Operational Space formulation \cite{Khatib1987} to underactuated constrained mechanical systems.
This new formulation is less efficient than \cite{Righetti2011} because it uses the inverse of the robot mass matrix.

These approaches based on the elimination of the contact forces present two major drawbacks.
First, in general they can not guarantee bounded contact forces.
Second, every time the robot makes or breaks a contact, the \emph{discontinuity} in the constraint set results in discontinuous control torques.
These discontinuities may generate jerky movements or, even worse, make the robot slip and fall.
\begin{figure}[!tbp]
   \centering
   \subfloat[]{ \includegraphics[height=3cm]{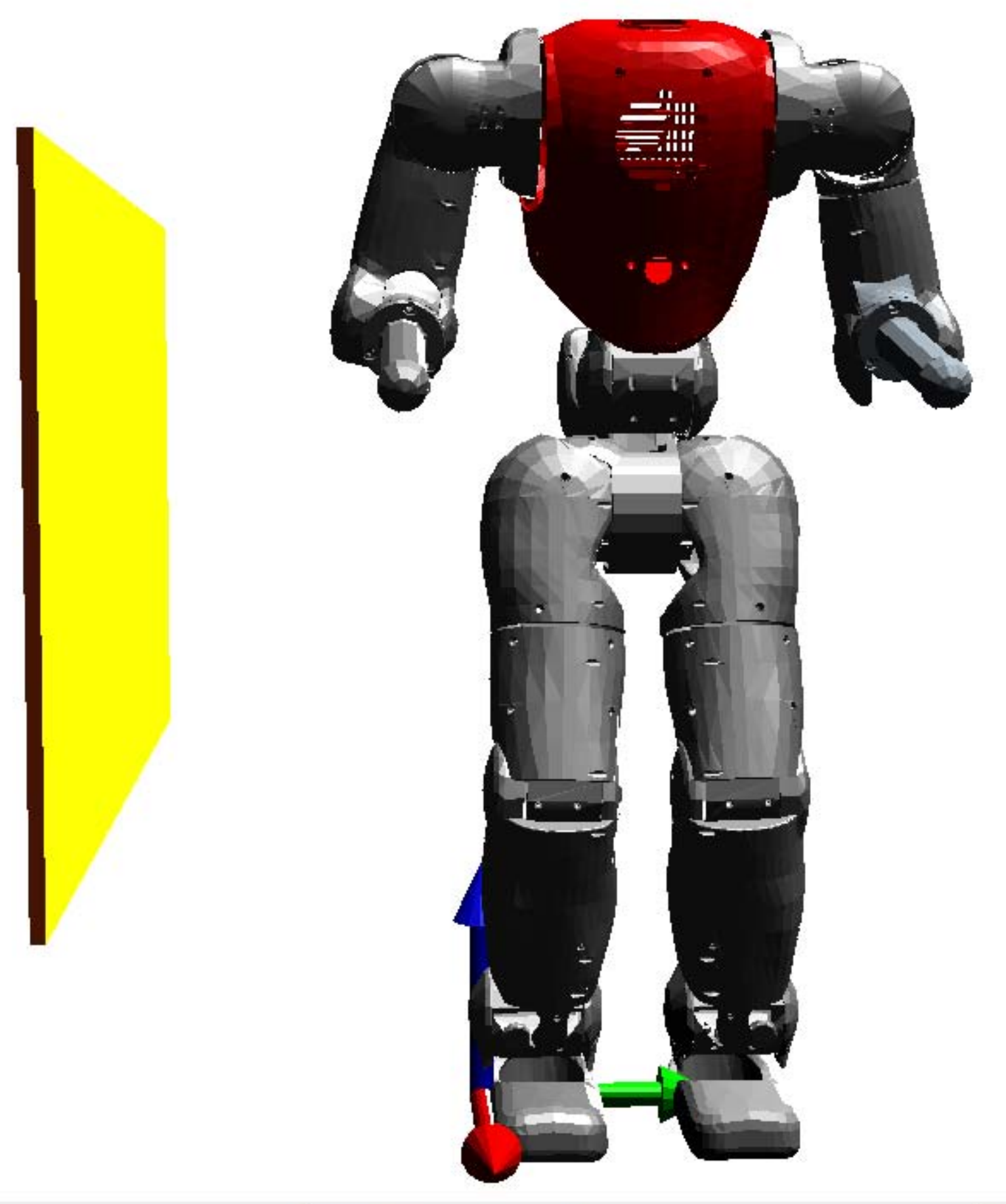}} \quad
   \subfloat[]{ \includegraphics[height=3cm]{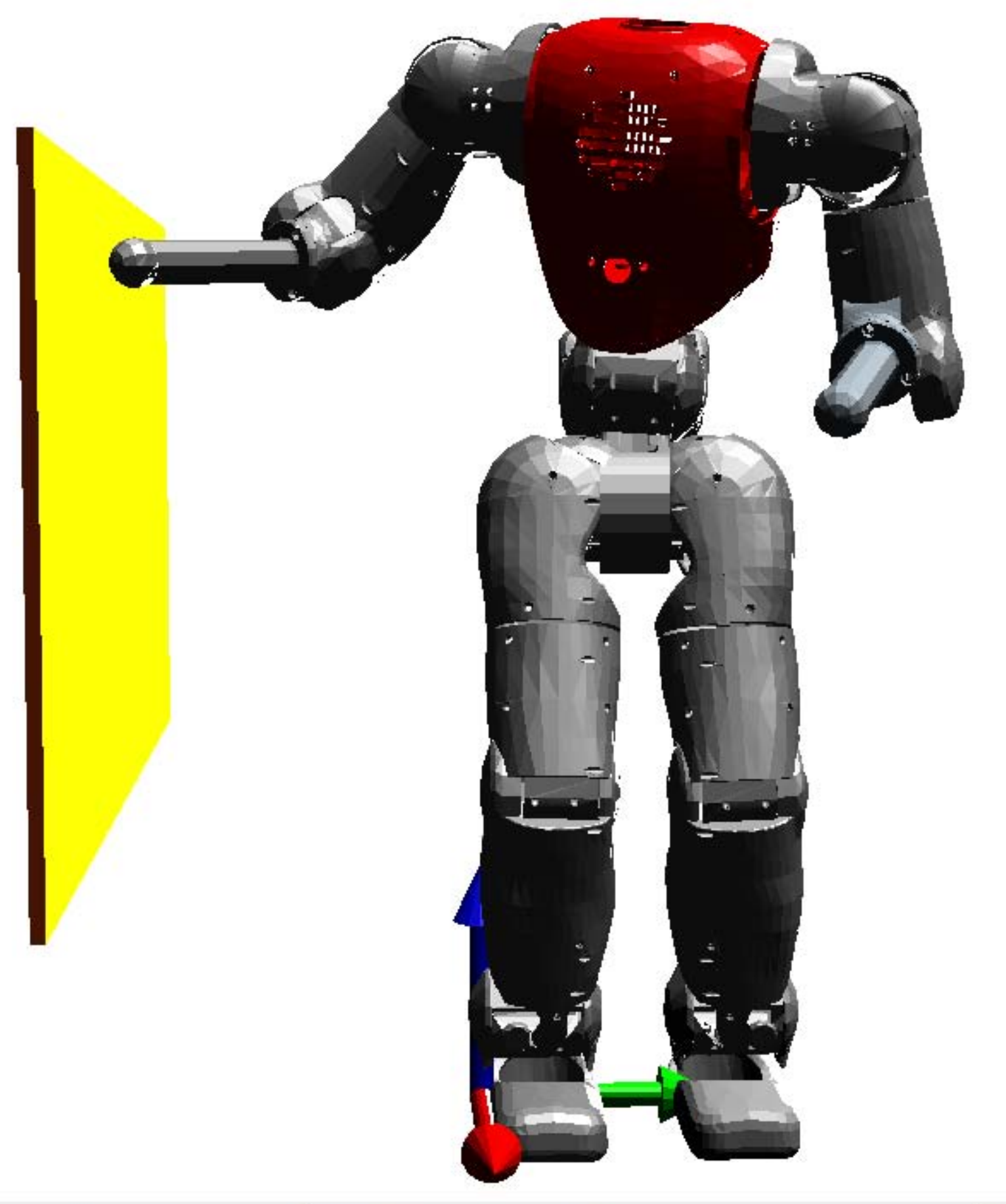}} \quad
   \subfloat[]{ \includegraphics[height=3cm]{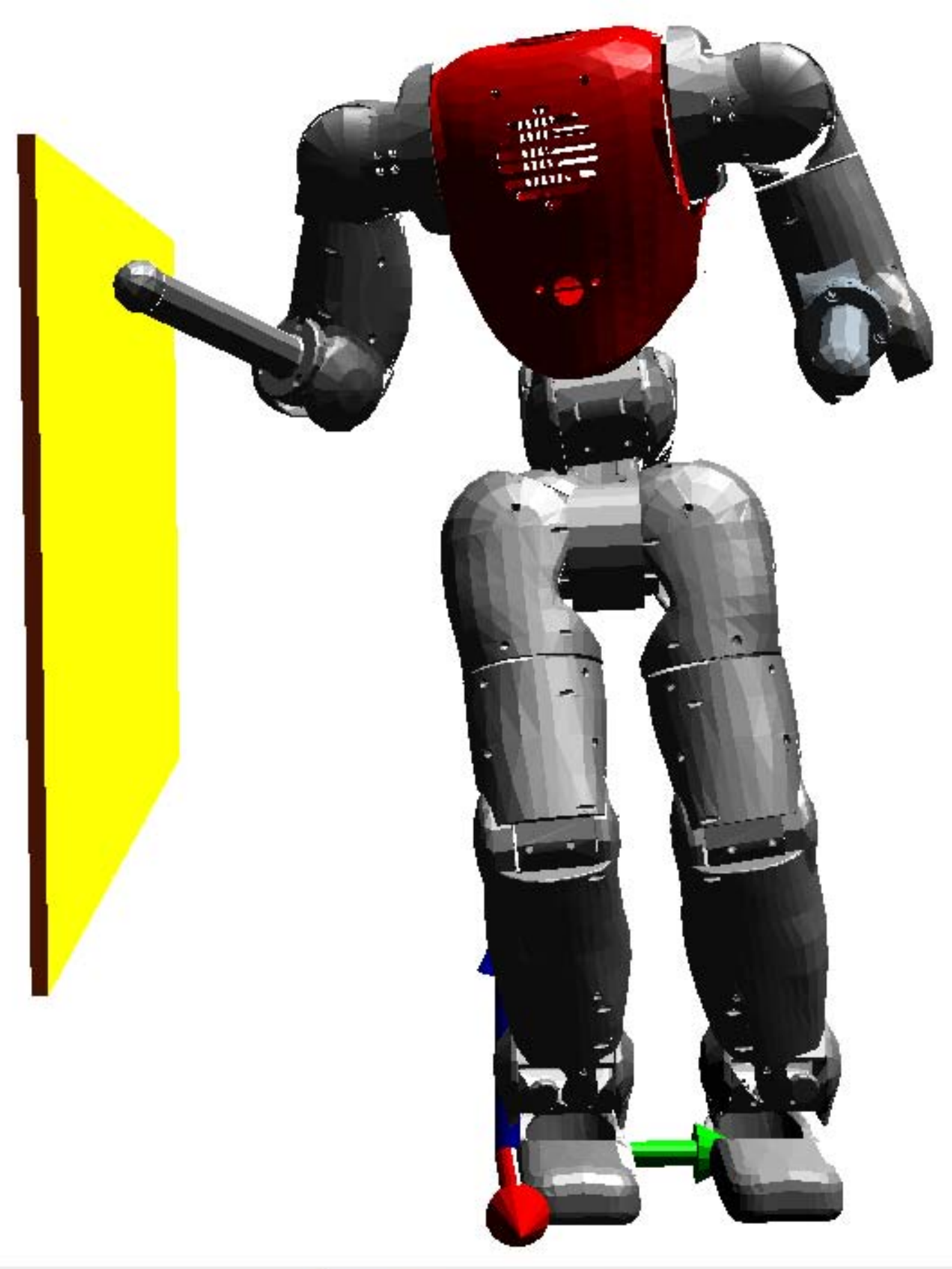}}\\
   \subfloat[]{ \includegraphics[height=3cm]{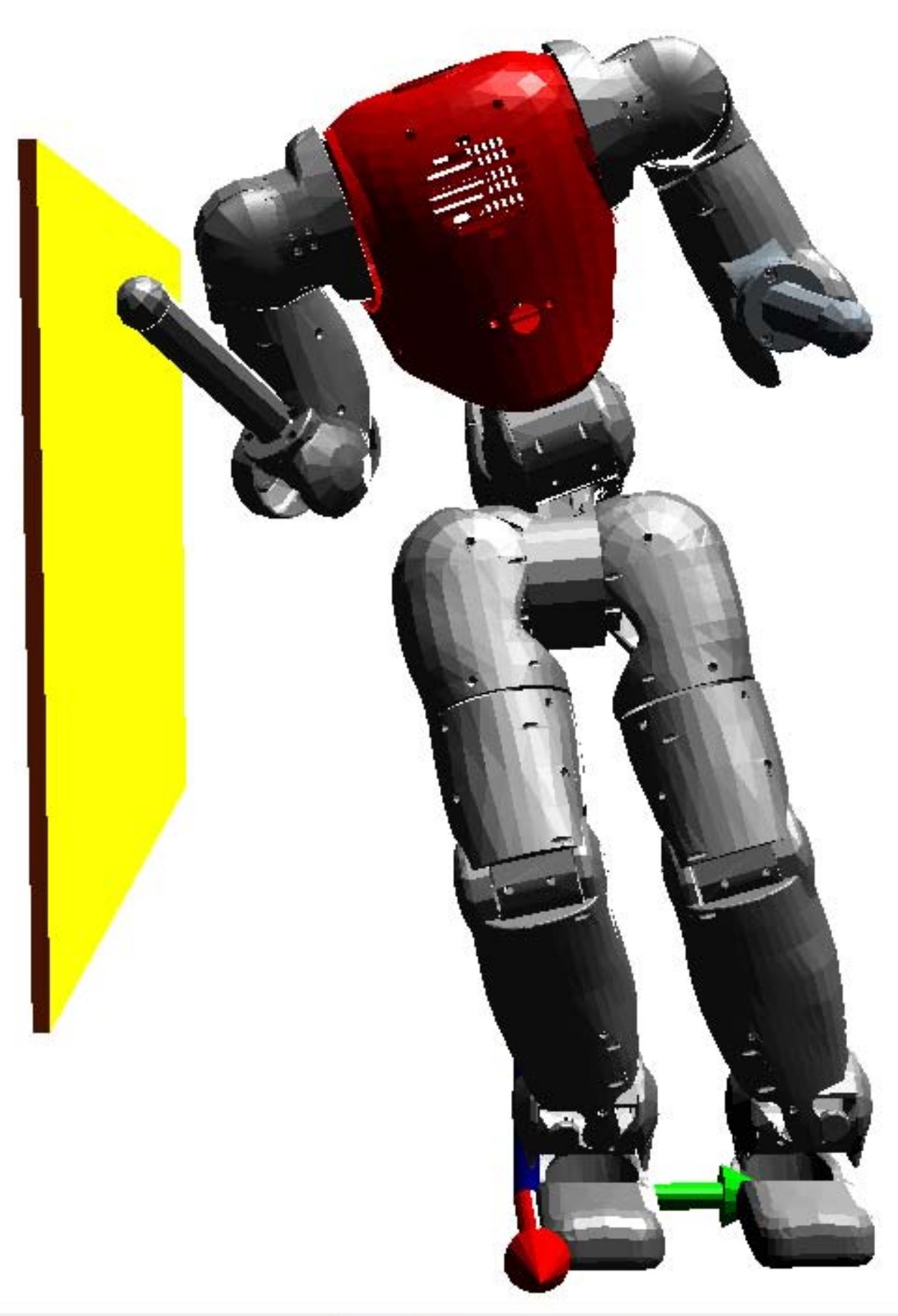}} \quad
   \subfloat[]{ \includegraphics[height=3cm]{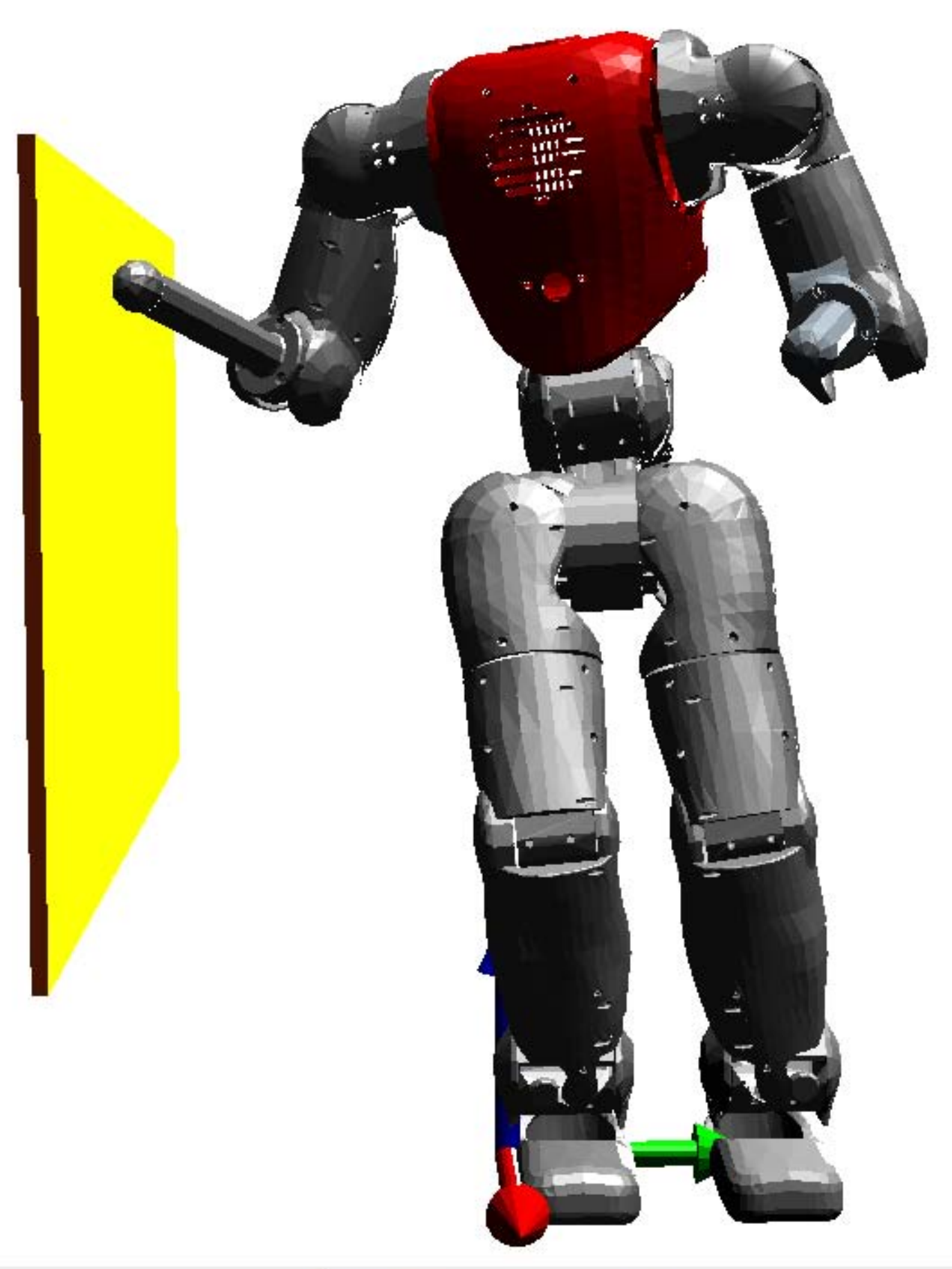}} \quad
   \subfloat[]{ \includegraphics[height=3cm]{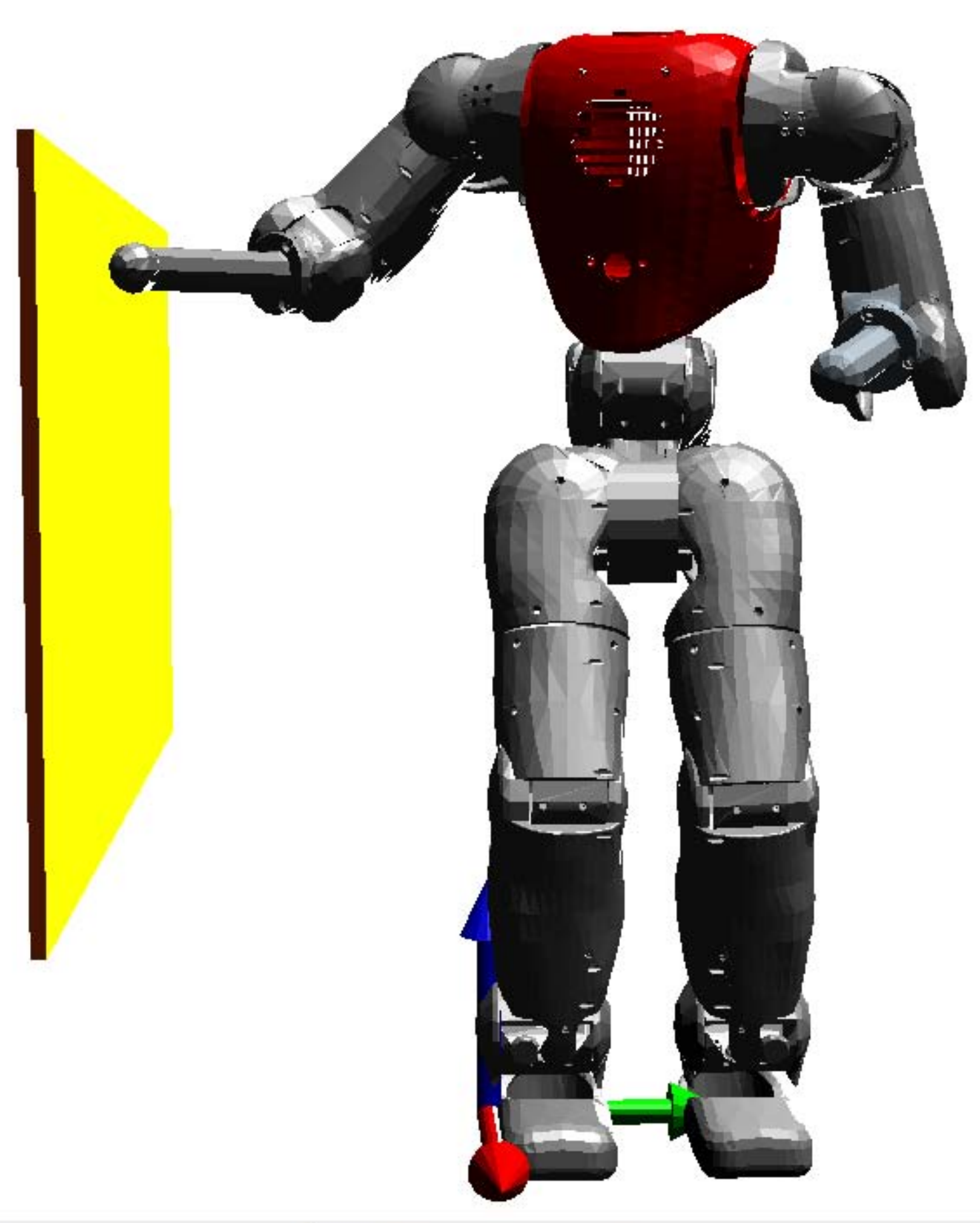}}
   \caption{Test 1. The robot made contact on the yellow wall; then it moved its COM towards the wall and back.}
   \label{fig:test1_screenshot}
\end{figure}

Rather than finding an analytical solution of the control problem, an alternative approach \cite{DeLasa2010} is to use a Quadratic Programming (QP) solver.
This allows to include inequality constraints into the problem formulation, which can model control tasks and physical constraints (e.g. joint limits, friction cones).
For instance, Saab et al. \cite{Saab2011} used inequalities to account for the Zero Moment Point (ZMP) conditions on a walking humanoid.
While this technique is appealing, solving a cascade of QPs with inequality constraints can be critical from a computational standpoint. 
Escande et al. \cite{Escande2014} reached a computation time of 1 ms on an inverse-kinematics problem --- at the price of seldom suboptimal solutions.
However, they did not consider the inverse-dynamics problem (as we do in this work), which has more than twice the number of variables and, consequently, is more computationally demanding.
In another recent work Herzog et al. \cite{Herzoga} succeeded in controlling their robot at 1 KHz using an inverse-dynamics formulation.
Nonetheless the robot had only 14 Degrees of Freedom (DoFs) and the CPU had 3.4 GHz; in case of more DoFs or slower CPU their method may still be too slow.

The main contribution of this paper is a convenient reformulation of the constrained optimization problem that arises in multi-task control frameworks such as \cite{DeLasa2010,Saab2011,Escande2014,Escande2010}.
We derive an analytical sparse solution of the problem constraints, which allows us to convert the original problem into two smaller independent unconstrained problems.
The resulting method has a computational cost similar to inverse-dynamics based methods \cite{Righetti2011}, while allowing for force control and presenting the flexibility of optimization-based techniques.
The paper is structured as follows.
Section \ref{sec:method} presents the theoretical results.
Section \ref{sec:tests} validates the presented control laws on a simulated 23-DoF humanoid robot.
Section \ref{sec:conclusions} summarizes the presented results and illustrates some future extensions.

\section{Method}
\label{sec:method}
This section introduces the analyzed control problem and motivates the need for a more efficient formulation.
Then we derive an analytical solution of the problem constraints, which allows us to simplify the optimization.
Finally we analyze the computational complexity of the new formulation, and we provide some insights into the physical principles that lie at the basis of our analytical work.

\subsection{Notation}
The state of a floating-base rigid robot with $n$ joints can be expressed as a vector $q\in \Rv{n+6}$, where the first 6 elements represent the position and orientation of the floating base (e.g. the hip link) and the remaining $n$ elements represent the joint angles. 
Suppose that the robot is subject to a set of $k$ nonlinear constraints: $e(q,\dot{q},t) = 0$, which for instance could be due to rigid contacts.
By differentiating the constraints (once or twice, depending on whether they are holonomic) we can express them at acceleration level.
We can then write the equations of motion of the system as:
\begin{subequations} \label{eq:constr_dyn}\begin{align} 
\label{eq:system_dyn} 
M(q) \ddot{q} + h(q,\dot{q}) - J_c(q)^\T f_c &= S^\T \tau \\
\label{eq:system_constr} 
J_c(q) \ddot{q} &= c_c(q,\dot{q},t),
\end{align}\end{subequations}
where 
$M \in \mathbb{R}^{(n+6) \times (n+6)}$ is the joint space mass matrix, 
\mbox{$\ddot{q} \in \mathbb{R}^{n+6}$} contains the joint and base accelerations, 
\mbox{$h \in \mathbb{R}^{n+6}$} contains the gravity, centrifugal and Coriolis forces, 
$S = \mat{0_{n\times 6} & I_{n\times n}} \in \mathbb{R}^{n\times (n+6)}$ is the joint selection matrix, 
$\tau \in \mathbb{R}^n$ are the joint torques, 
$J_c = \pder{e}{q} \in \mathbb{R}^{k \times (n+6)}$ is the constraint Jacobian,  
$f_c \in \mathbb{R}^{k}$ are the constraint forces and
$c_c \in \Rv{k}$ is a term resulting from the derivation of the nonlinear constraints $e(q,\dot{q},t)$.
We can rewrite \eqref{eq:system_dyn} and \eqref{eq:system_constr} as a unique affine function $Dy=d$ of the variable $y^\T = \mat{\ddot{q}^\T & f_c^\T & \tau^\T}$.
We now split the constraints into two subsets: the \emph{controlled constraints} (with Jacobian $J_f \in \mathbb{R}^{k_f\times (n+6)}$ and forces $f_f \in \mathbb{R}^{k_f}$), and the \emph{supporting constraints} (with Jacobian $J_s\in \mathbb{R}^{k_s\times n+6}$ and forces $f_s \in \mathbb{R}^{k_s}$), so that:
\begin{equation*}
J_c^\T = \mat{J_f^\T & J_s^\T}, \quad f_c^\T = \mat{f_f^\T & f_s^\T}, \quad c_c^\T = \mat{c_f^\T & c_s^\T}
\end{equation*}
This division is motivated by the fact that we mean to directly control $f_f$, while we use $f_s$ only to support the system.
We consider that $J_f$ and $J_s$ may be rank deficient, i.e. $\rank{J_f} = \hat{k}_f \neq k_f$, $\rank{J_s} = \hat{k}_s \neq k_s$, but we assume that the two sets of constraints are linearly independent, i.e. $\rank{J_c} = \hat{k} = \hat{k}_f + \hat{k}_s$.
We finally introduce our notation for the basis matrices with an example: we represent with $Z_s \in \mathbb{R}^{(n+6) \times (n+6-\hat{k}_s)}$ an orthonormal basis of the \emph{nullspace} of $J_s$, and with $U_s \in \mathbb{R}^{(n+6) \times \hat{k}_s}$ an orthonormal basis of the \emph{range} of $J_s$.
Similarly, the basis matrices of $J_f$, $J_c$, $J_f Z_s$ and $Z_s S^\T$ are denoted by the subscripts $f$, $c$, $fs$ and $ss$, respectively.

\subsection{Multi-Task Motion and Force Control}
We consider an arbitrary number $N$ of control tasks that can be represented as convex quadratic functions \mbox{$g_i(y) = ||A_i y - a_i ||^2$} --- in particular, functions of $\ddot{q}$ and $f_f$.
Moreover, we suppose that tasks have different priorities, that is, in case of conflict, tasks with higher priority should be satisfied at the expenses of the tasks with lower priority.
We can then formulate the multi-task control problem as a cascade of constrained optimizations \cite{Escande2014}:
\begin{equation} \label{eq:ctrl_prob1} \begin{split}
\lexmin_{y \in \mathbb{R}^{2n+k+6}}\,& \{g_1(y), \dots, g_N(y)\}\\
s.t. \quad & Dy=d \\
\end{split} \end{equation}
where, at the optimum, each cost function $g_i(y)$ is minimized with respect to a \emph{lexicographic} order: 
it is not possible to decrease an objective $g_i$ without increasing an objective $g_j$ with higher priority (i.e. $j<i$).
Once we have found the solution $y^*$, we can command to the motors the joint torques contained in it.

Problem \eqref{eq:ctrl_prob1} is a generic formulation of the operational-space inverse-dynamics problem. 
This formulation consists of $N$ sequential optimizations, each with $2n+k+6$ variables and $n+6+k$ equality constraints.
While we could use a generic QP solver to compute the lexicographic optimum \cite{Saab2013, Mansard2012}, 
the structure of the problem has a specific shape that we can use to simplify the computation. 
Some parts of the structure was used in \cite{Sentisa, Mistry2011}, with some specific hypothesis. 
We propose here a more generic though more efficient formulation to fully exploit the problem sparsity and the algorithmic structure.

We start by considering that all the solutions of \eqref{eq:constr_dyn} take the following form:
\begin{equation} \label{eq:constr_sol}
y = y^* + K z_D,
\end{equation}
where $y^*$ is such that $Dy^*=d$, the columns of $K$ span the nullspace of $D$ and $z_D$ is a free parameter.
A numerical solver typically computes $y^*$ and $K$ through a decomposition\footnote{The classical nullspace approach uses $y^*=D^\pinv d$ and $K=Z_D$, i.e. an orthogonal basis of the nullspace of $D$} of the matrix $D$, substitutes \eqref{eq:constr_sol} inside the cost functions of \eqref{eq:ctrl_prob1} and solves the resulting unconstrained problem with variable $z_D \in \Rv{n+k-\hat{k}}$ \cite{DeLasa2010}.
However, the decomposition of $D$ is costly and, in general, this numerical approach results in a dense $K$.
We find an analytical expression of the solutions of \eqref{eq:constr_dyn}, so we do not need to decompose $D$.
The analytical solution results in a sparse $K$, which allows us to reformulate \eqref{eq:ctrl_prob1} as two independent unconstrained optimizations. 
Moreover, the proposed formulation does not require the computation of the mass matrix $M$.

\subsection{Analytical Sparse Solution of the Constraints}
The solution that we are about to derive builds on the assumption that the mechanical system is \emph{sufficiently constrained}.
\begin{definition}
We say that a constrained mechanical system is \emph{sufficiently constrained} if the Jacobian of the supporting constraints 
satisfies this condition:
\begin{equation}\label{eq:suff_constr}
\rank{J_s \bar{S}^\T} = 6,
\end{equation}
where $\bar{S} = \mat{I_{6\times 6} & O_{6\times n}}$.
\end{definition}
To solve \eqref{eq:constr_dyn} we start by solving \eqref{eq:system_constr}:
\begin{equation} \label{eq:acc_solution}
\ddot{q} = J_c^\pinv c_c + Z_c z_c,
\end{equation}
where $z_c\in \Rv{n+6-\hat{k}}$ is a free parameter.
Now we substitute \eqref{eq:acc_solution} in \eqref{eq:system_dyn} and we project the resulting equation in the nullspace of the supporting constraints:
\begin{equation} \label{eq:sup_proj_dyn}
Z_s^\T (M (J_c^\pinv + Z_c z_c) + h - J_f^\T f_f) = Z_s^\T S^\T \tau
\end{equation}
Since the system is \emph{sufficiently constrained} $Z_s^\T S^\T$ is full-row rank (see \cite{DelPrete2014_app} for the proof), hence for any value of $f_f$ and $z_c$ we can find a value of $\tau$ that satisfies \eqref{eq:sup_proj_dyn}, that is:
\begin{equation} \label{eq:tau_solution}
\tau = (Z_s^\T S^\T)^\pinv Z_s^\T (M (J_c^\pinv + Z_c z_c) + h - J_f^\T f_f) + Z_{ss} z_{ss},
\end{equation}
where $z_{ss} \in \Rv{\hat{k}_s-6}$ is a free parameter.
Finally, since $J_f$ may be rank deficient, we switch to a minimal representation of $f_f$ in terms of a free parameter $z_f\in \Rv{\hat{k}_f}$:
\begin{equation} \label{eq:ff_solution}
f_f = U_f z_f + (I - U_f U_f^\T) \hat{f}_f,
\end{equation}
where $\hat{f}_f$ is a measurement of $f_f$, which is necessary only if $J_f$ is rank deficient.
Using \eqref{eq:acc_solution}, \eqref{eq:tau_solution} and \eqref{eq:ff_solution} we can write:
\begin{equation} \label{eq:analytic_sparse_sol}\begin{split}
\mat{\ddot{q} \\ f_f  \\ f_s \\ \tau} &= \mat{\ddot{q}^* \\ f_f^*  \\ f_s^* \\ \tau^*} +
\mat{  K_{cc}      & 0                 &    0        \\
          0               & K_{ff}              &    0         \\
          K_{sc}       & K_{sf}         &  K_{s\tau} \\
          K_{\tau c}  & K_{\tau f}   & K_{\tau \tau}   }
\mat{z_c \\ z_f \\ z_{ss}}, \\
\end{split}\end{equation}
with:
\begin{equation*} \begin{split}
\ddot{q}^*   &= J_c^\pinv c_c, \qquad
f_f^*             = (I - U_f U_f^\T) \hat{f}_f \\
\tau^*         &= (Z_s S^\T)^\pinv Z_s (MJ_c^\pinv c_c + h -J_f^\T f_f^*) \\
K_{cc}        &= Z_c, \qquad
K_{ff}            = U_f, \qquad
K_{\tau \tau} = Z_{ss} \\
K_{\tau c}   &= (Z_s S^\T)^\pinv Z_s M Z_c \\
K_{\tau f}    &= -(Z_s S^\T)^\pinv Z_s  J_f^\T U_f
\end{split}\end{equation*}
We do not report here the values of $K_{sc}, K_{sf}, K_{s\tau}$ and $f_s^*$ because we do not use them in our formulation.
Finally, \eqref{eq:analytic_sparse_sol} is a sparse analytical representation of the solutions of \eqref{eq:constr_dyn}.

\subsection{New Problem Formulation}
Using \eqref{eq:analytic_sparse_sol} we can now express problem \eqref{eq:ctrl_prob1} in terms of the new variables $z_c, z_f, z_{ss}$.
A clear decoupling appears in \eqref{eq:analytic_sparse_sol}: $\ddot{q}$ (the motion) only depends on $z_c$, while $f_f$ (the force) only depends on $z_f$.
Since by assumption the tasks $g_i(y)$ only depend on $\ddot{q}$ and $f_f$, we can exploit this decoupling.
Without loss of generality we assume that each control task is function of either $f_f$ or $\ddot{q}$ (if not, we can split the task into two separate tasks with arbitrary order).
The task matrices $A_i$ have then the following structure \mbox{$A_i = \mat{A^q_i & A^f_i & 0_{m_i\x k_s+n}}$},
and we define $\mathcal{I}_f$ and $\mathcal{I}_q$ as the set of indexes of the force tasks and the motion tasks, respectively.

Under these conditions, the solutions of \eqref{eq:ctrl_prob1} can be computed through \eqref{eq:tau_solution}, where $z_{ss} \in \Rv{\hat{k}_s-6}$ is an arbitrary vector, whereas \mbox{$z_f\in \Rv{\hat{k}_f}$} and $z_c\in \Rv{n+6-\hat{k}}$ are the solutions of two independent lexicographic optimizations.
The first optimization finds the desired constraint forces by minimizing the cost functions:
\begin{equation} \label{eq:force_hier}
g_i(z_f) = ||A_i^f U_f z_f - a_i + A_i^f (I - U_f U_f^\T) \hat{f}_f||^2, \quad \forall i \in \mathcal{I}_f
\end{equation}
The second optimization finds the desired joint accelerations by minimizing:
\begin{equation} \label{eq:motion_hier}
g_i(z_c) = ||A^q_i Z_c z_c + A^q_i J_c^\pinv c_c - a_i ||^2, \qquad \forall i \in \mathcal{I}_q
\end{equation}

\subsection{Computational Complexity}
\label{sec:compu_compl}
Even though the new optimizations \eqref{eq:force_hier} and \eqref{eq:motion_hier} have less variables and constraints, we must consider the cost incurred in reformulating the problem. 
This cost is dominated by the computation of the following five matrices:
\begin{equation*}
Z_s, \spac U_f, \spac Z_c,\spac J_c^\pinv,\spac (Z_s^\T S^\T)^\pinv Z_s^\T
\end{equation*}
We can get the first two matrices by computing an SVD \footnote{Alternatively we can use any other complete rank-revealing decomposition, e.g. the Complete Orthogonal Decomposition \cite{Golub1996}} of $J_s$ and $J_f$.
Then, to compute $Z_c$ and $J_c^\pinv$ we only need to decompose $J_f Z_s$ and exploit the following relationships:
\begin{equation*}
J_c^+ = \mat{ Z_s(J_f Z_s)^\pinv & (I-Z_s(J_f Z_s)^\pinv J_f) J_s^\pinv }, \quad Z_c = Z_{fs} Z_s
\end{equation*}
Finally, thanks to the assumption \eqref{eq:suff_constr}, we only need to decompose $\bar{S} J_s^\T$ to compute the last matrix:
\begin{align*}
(Z_s^\T S^\T)^\pinv Z_s^\T &= \mat{ -S J_s^\T (\bar{S} J_s^\T)^\pinv & I }
\end{align*}
Considering that an SVD of an $m\x n$ matrix (with \mbox{$m<n$}) has a cost $O(m^2n)$, and that typically $n>k$, the total expected cost for decomposing these four matrices is: 
\begin{equation*}
O((n+6) (k_s^2 + k_f^2) + (n+6-\hat{k}_s) k_f^2 + 36 k_s) = O(n (k_s^2+2k_f^2))
\end{equation*}
Conversely, decomposing the constraint matrix $D$ of \eqref{eq:ctrl_prob1} has a cost $O((n+k)^2 (2n+k))$.
We can gather that the cost of our formulation is always less than the cost of resolution of the original constraints.

Moreover, our formulation has two additional advantages. 
First, there is no need to compute the mass matrix of the robot $M$ because we can compute \eqref{eq:tau_solution} using the Recursive Newton-Euler Algorithm (RNEA) \cite{featherstone2008rigid}.
Second, the force and the motion hierarchies are independent, hence they can be solved in parallel.

\subsection{Physical Interpretation}
The condition \eqref{eq:suff_constr} has been erroneously approximated in previous works with the less strict condition $\rank{J_s}\ge 6$.
Actually it is true that \eqref{eq:suff_constr} implies that $\rank{J_s}\ge 6$, but not the opposite --- for instance, a point-foot quadruped with two feet on the ground verifies the second condition, but not the first one.
The intuitive reason why we need at least six independent constraint forces that we are willing not to control is that these constraints compensate for the 6 degrees of underactuation of the system. 
In the constraint-consistent space the system is then fully-actuated, because \emph{the supporting constraint forces can accelerate the floating base in all six directions}.
In practice, a humanoid robot standing with (at least) one foot flat on the ground always satisfies this condition --- in fact we can see it as a fixed-base manipulator.
This allows us to feedback-linearize the system and decouple kinematics and dynamics. 

When the robot is not \emph{sufficiently constrained} (i.e. \eqref{eq:suff_constr} is not satisfied), we cannot apply the proposed formulation as it is.
In that situation the system loses the complete control over its momentum, hence we can no longer decouple kinematics and dynamics.
Nonetheless, we can use the same insights to find another convenient formulation for that case.
This is subject of ongoing work.

\section{Tests}
\label{sec:tests}
This section presents two simulation tests that validate our control framework and demonstrate its potential and benefits.

\subsection{Experimental Setup}
We carried out the tests on a customized version of the Compliant huManoid (CoMan) simulator \cite{Dallali}.
Table~\ref{table:sim_param} lists the parameters of the simulation environment.
\begin{table}[!hbt] 
\caption{Simulation parameters.}
\centering 
\begin{tabular}{p{2.2cm} p{1.5cm} | p{2.1cm} p{1.1cm}} 
\hline 
      	 Contact stiffness  			&    $2 \cdot10^5 N/m$ 				&   Contact damping 				&    $10^3 Ns/m$\\ \rowcolor[gray]{.9}
	 Integration relative tolerance	&	$10^{-3}$						&   Integration absolute tolerance	&  $10^{-6}$ 	\\ 
	 Integration scheme			&	\emph{ode23t} \cite{matlabOde23t}	&   Robot DoFs					&  23+6 	\\ \rowcolor[gray]{.9}
	 Control frequency			& 	1 KHz						&   CPU						& 2.83 GHz \\
[0.5ex] \hline 
\end{tabular} 
\label{table:sim_param} 
\end{table}

\subsubsection{Motion Control}
To control an operational point \mbox{$x(q) \in \Rv{m}$} of the robot we use the kinematic relationship:
\begin{equation*}
J \ddot{q} = \ddot{x} - \dot{J} \dot{q},
\end{equation*}
where $J \in \R{m}{n+6}$ is the Jacobian associated to $x$.
Since the presented control framework works at acceleration level, a drift is likely to occur.
To prevent deviations from the desired trajectory and to ensure disturbance rejection, we computed the desired task accelerations $\ddot{x}^* \in \Rv{m}$ with a proportional-derivative feedback control law:
\begin{equation*}
\ddot{x}^* = \ddot{x}_r + K_d (\dot{x}_r - \dot{x}) + K_p (x_r - x),
\end{equation*}
where $x_r(t), \dot{x}_r(t), \ddot{x}_r(t) \in \Rv{m}$ are the position-velocity-acceleration reference trajectories, whereas $K_d \in \R{m}{m}$ and $K_p\in \R{m}{m}$ are the diagonal positive-definite matrices.
To generate $x_r(t), \dot{x}_r(t), \ddot{x}_r(t)$ we used the approach presented in \cite{Pattacini2010}, which provides approximately minimum-jerk trajectories.
We set all the proportional gains $K_p=10 s^{-2}$, and all the derivative gains $K_d=5 s^{-1}$.

\subsubsection{Supporting Force Optimization}
\label{sec:force_opt}
When the matrix $Z_s^\T S^\T$ has a nontrivial nullspace, there are infinite joint torques that generate the same controlled forces and joint accelerations:
\begin{align*} 
\tau_1 =& M (J_c^\pinv + Z_c z_c) + h -J_f^\T U_f z_f \\
\tau =& (Z_s^\T S^\T)^{\pinv_W} Z_s^\T \tau_1 + Z_{ss} Z_{ss}^\T \tau_0,
\end{align*}
where $(.)^{\pinv_W} = W^{\half} (. W^{\half})^\pinv$ indicates a weighted pseudoinverse, with $W \in \R{n}{n}$ being an arbitrary positive-definite matrix and $\tau_0 \in \Rv{n}$ an arbitrary vector.
Any secondary objective can be considered by selecting arbitrary $W$ and $\tau_0$. 
Following the approach of Righetti et al. \cite{Righetti2013}, in our tests we have chosen to minimize a cost of the form $||f_s^\T W_f^{-1} f_s||^2$. 
This is achieved by setting:
\begin{align} \begin{split} \label{eq:force_opt}
W        =& (S(J_s^\pinv W_f^{-1} J_s^{\T \pinv} + Z_s Z_s^\T)S^\T)^{-1}\\
\tau_0 =& W S (J_s^\pinv W_f^{-1} J_s^{\T \pinv} + Z_s Z_s^\T) \tau_1
\end{split} \end{align}
The generalized inverse weighted by $W$ in the previous equation can be brought back to the pseudoinverse solution \eqref{eq:sup_proj_dyn} using the fact that 
\mbox{$A^{\pinv_W} = (I - Z_A (W Z_A)^\pinv W) A^\pinv$} \cite{Ben-Israel2003}. 
Even with this extension we do not need $M$ to compute $\tau$, but we can just use the RNEA.

\subsection{Computation Times}
To understand the practical implications of the proposed formulation we carried out a test for a typical case: \mbox{$n=23$}, $k_s=6$, $k_f=12$.
We measured the computation time taken to convert the original constrained problem into an unconstrained optimization.
With our approach, the most expensive operations in this phase are the four matrix decompositions discussed in Section~\ref{sec:compu_compl}.
Using the linear algebra C++ library \emph{Eigen}\cite{eigen}, we measured an average time of 0.23 ms for computing the four SVDs. 
Conversely, with the standard numerical approach, we measured an average time of 4.3 ms for decomposing the matrix $D$.
Considering that high-performance control loops require computation times below 1 ms, the observed $19\x$ speed-up could be critical for implementation on a real platform.

\subsection{Test 1 - Multi-contact force control}
In this test the robot made contact with a rigid wall using its right hand (see Fig.~\ref{fig:test1_screenshot}), and it regulated the contact force along the wall normal direction to 20 N.
The contact forces at the feet were considered as \emph{supporting forces}, so they were not controlled.
After making contact, we shifted the desired position of the Center Of Mass (COM) towards the right foot of the robot (i.e. along the y direction), so that the robot leaned against the wall, exploiting the additional support provided by the contact on its hand.
We report here the overall control hierarchy, in priority order: 
\begin{itemize}
\item constraints, both feet (12 DoFs);
\item force control, right hand (1 DoF); 
\item position control, COM ground projection (2 DoFs); 
\item position control, posture (29 DoFs).
\end{itemize}
The Root Mean Square Error (RMSE) for the force task was about 0.01 N, while for the COM task it was about 0.6 mm.
This kind of behavior is difficult to achieve with previous techniques \cite{Righetti2011,Sentis2008} that were mainly designed for locomotion and do not allow for direct control of interaction forces.

\subsection{Test 2 - Walking}
\begin{figure}[!tbp]
   \centering
   \subfloat[0 s: COM in the middle.]{ \includegraphics[height=3cm]{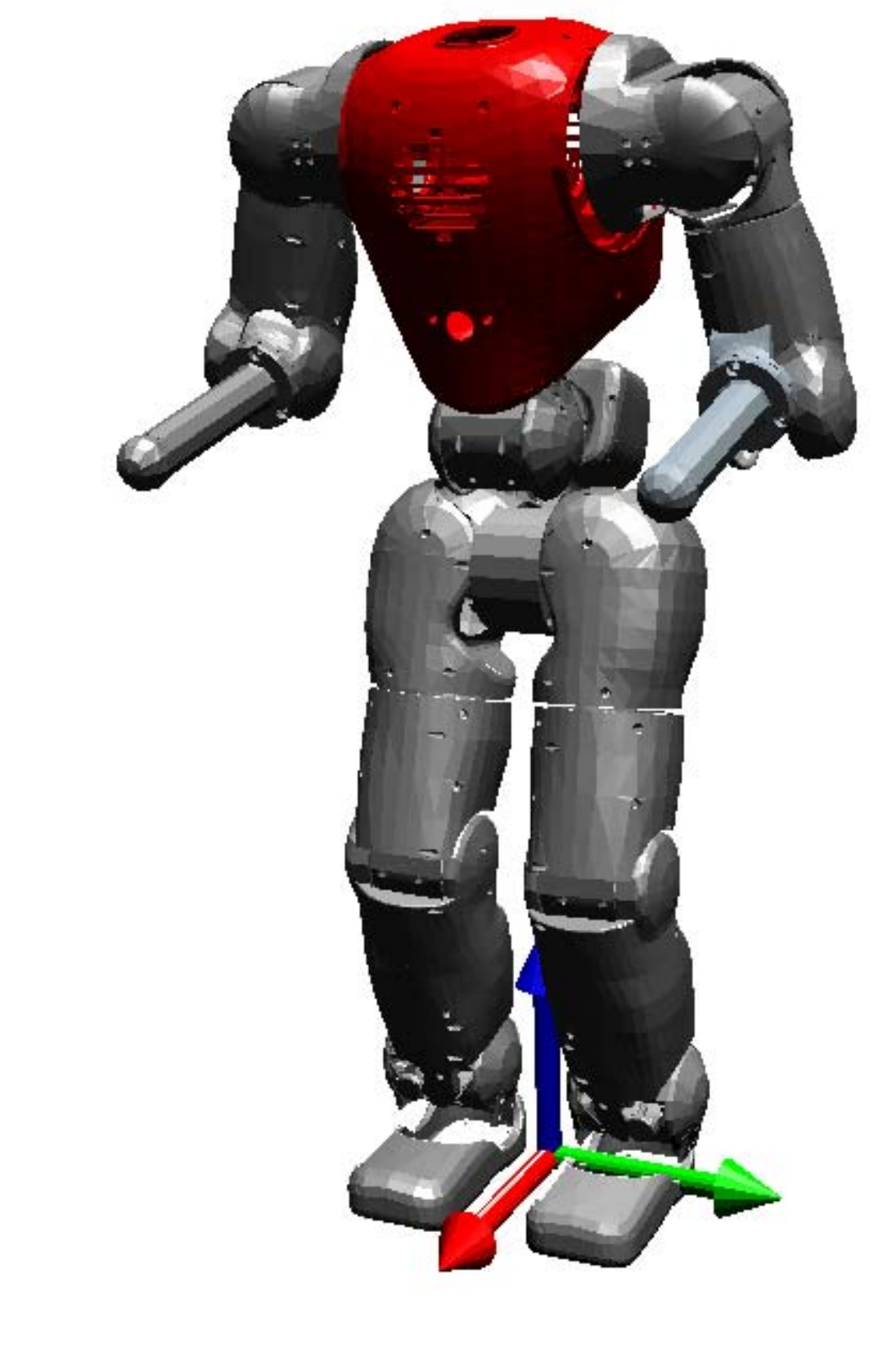}} \qquad
   \subfloat[2 s: COM over left foot, about to lift right foot.]{ \includegraphics[height=3cm]{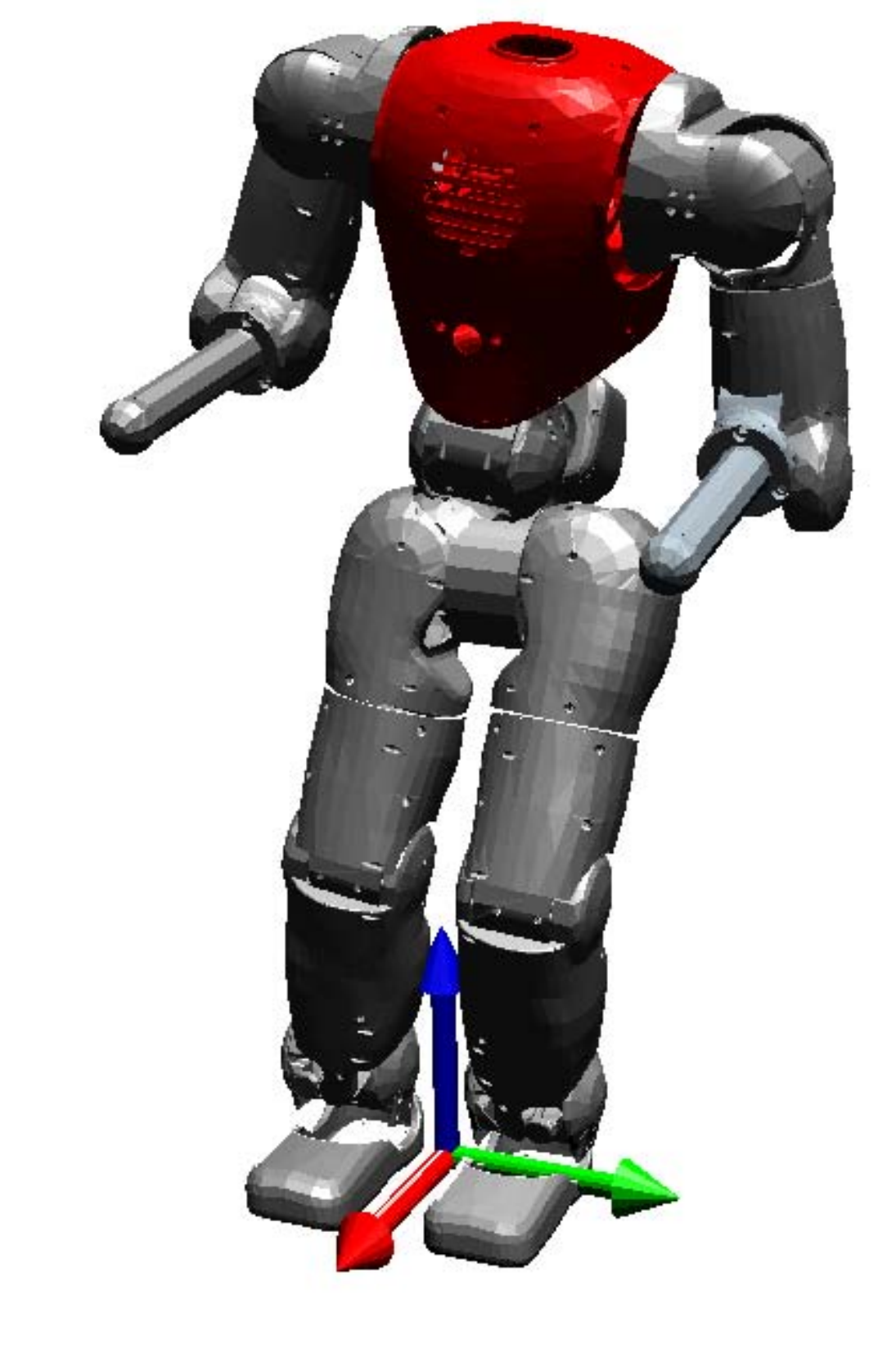}}\qquad
   \subfloat[3 s: stepping forward.]{ \includegraphics[height=3cm]{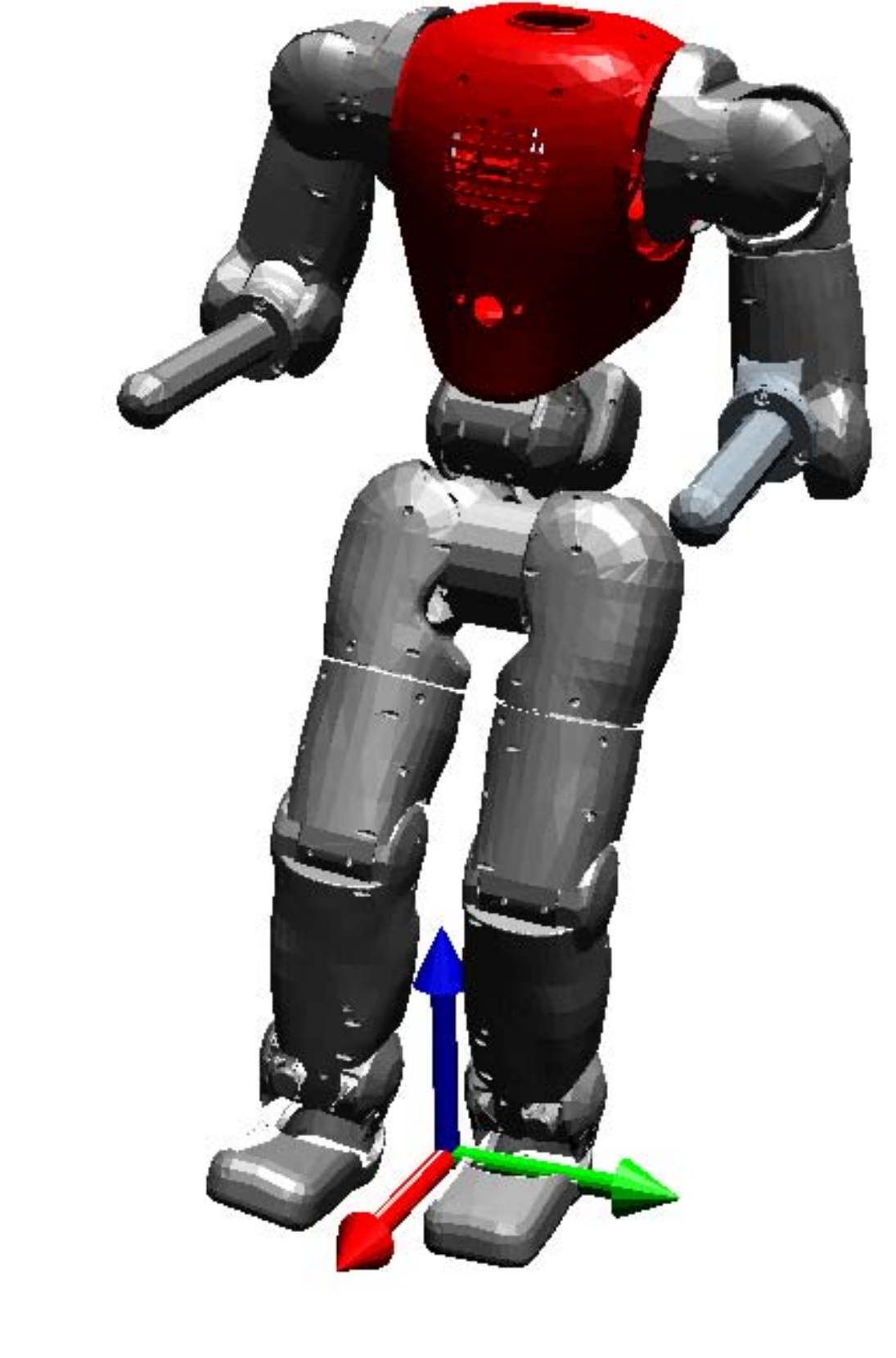}}\\
   \subfloat[4 s: ground impact.]{ \includegraphics[height=3cm]{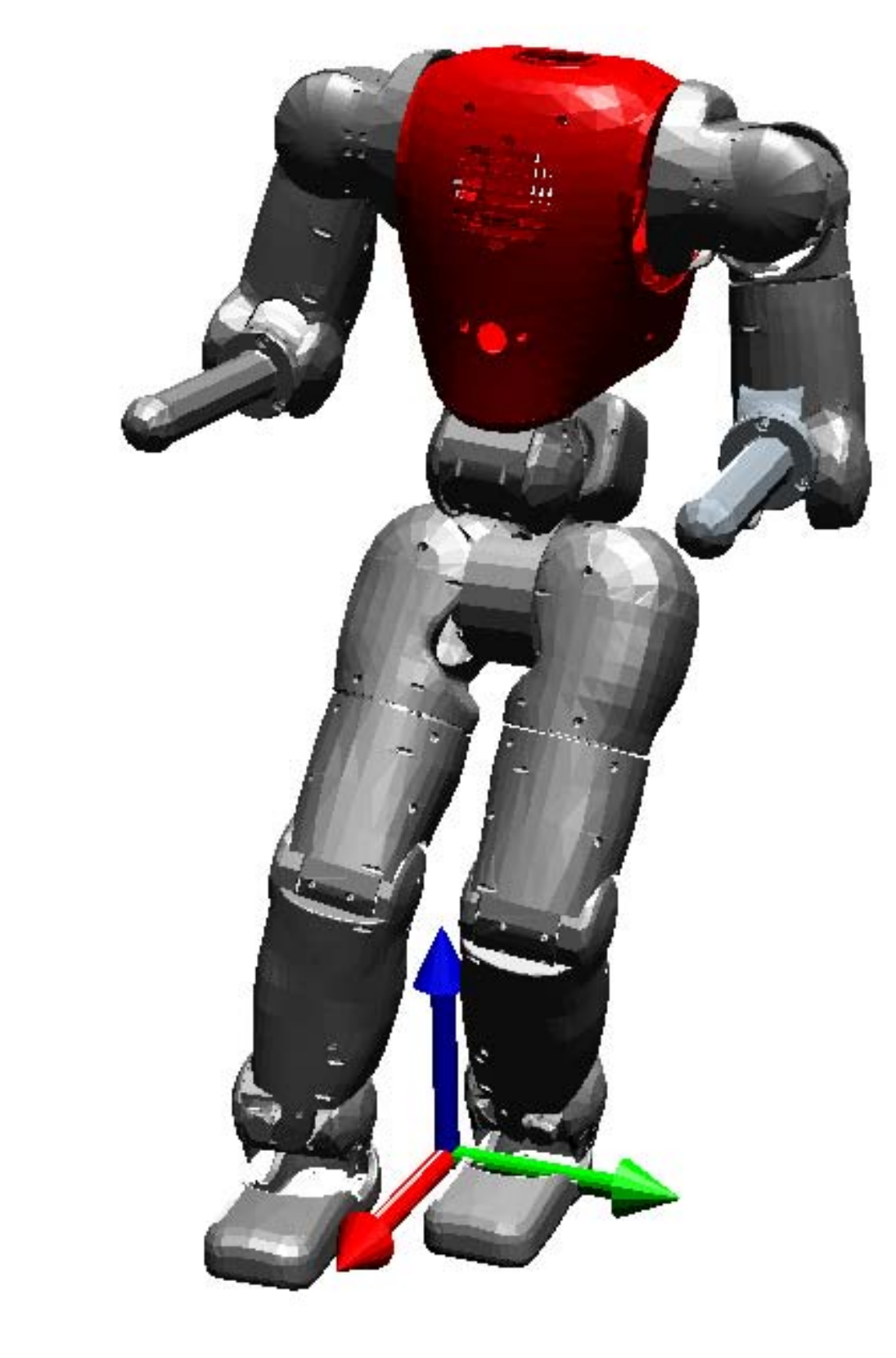}}\qquad
   \subfloat[6 s: COM in the middle.]{ \includegraphics[height=3cm]{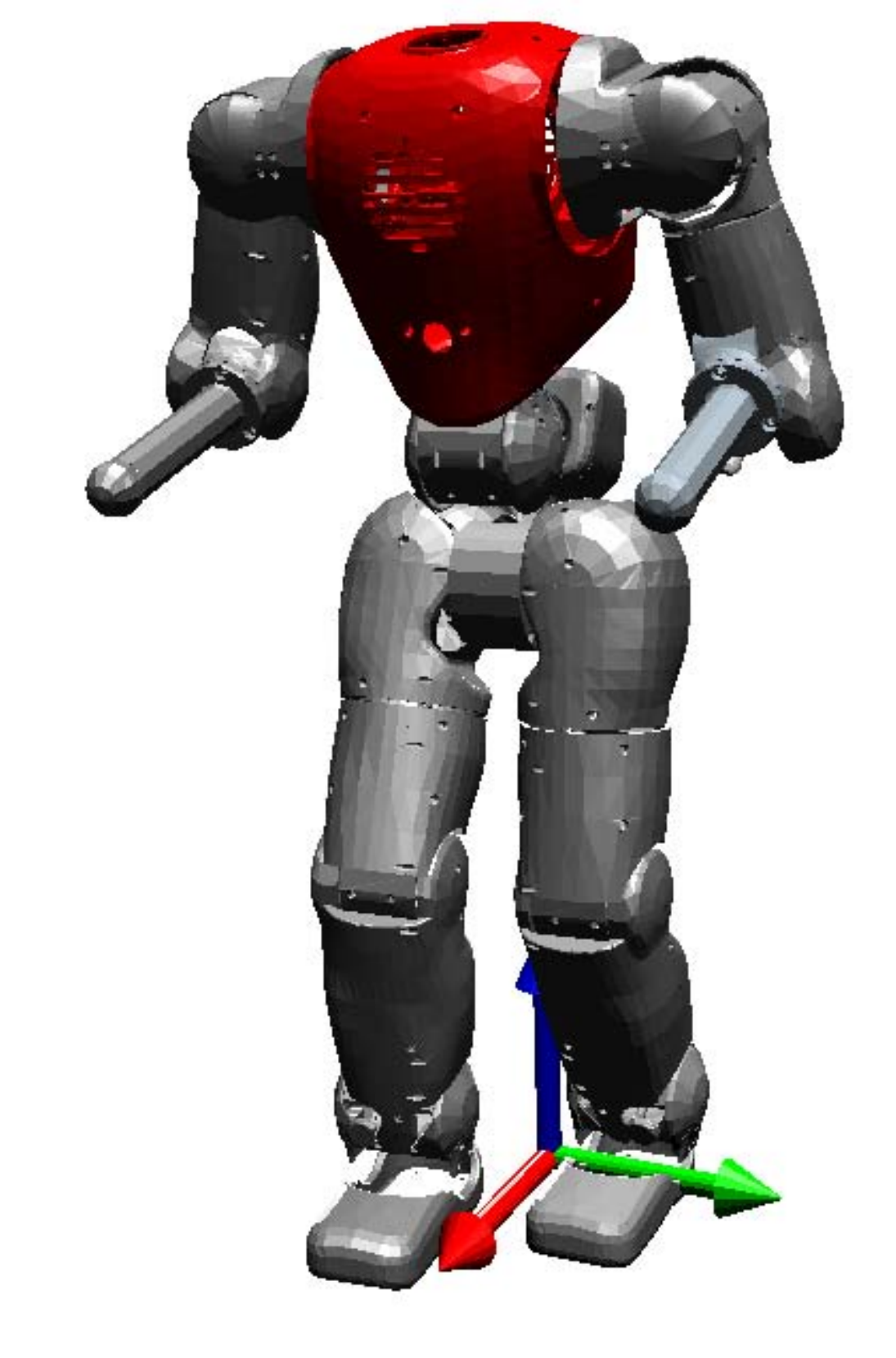}}\qquad
   \subfloat[8 s: COM over right foot.]{ \includegraphics[height=3cm]{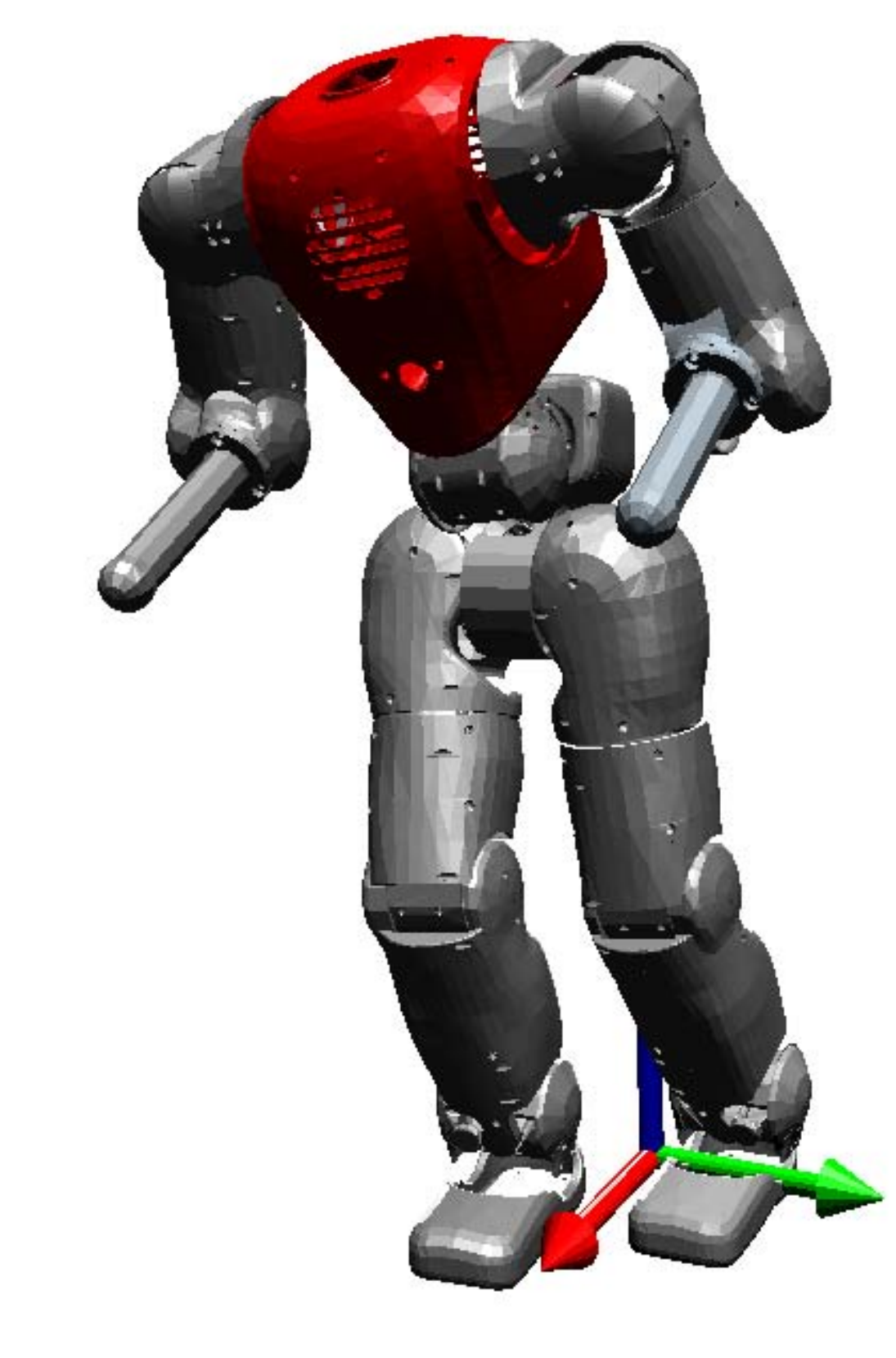}}
   \caption{Test 2. The robot performing a step with its right foot.}
   \label{fig:test2_screenshots}
\end{figure}

This test tackles the switching between different constraint phases, which, for instance, occurs when moving from single to double support during walking (see Fig.~\ref{fig:test2_screenshots}).
These hard constraint switches cause discontinuities in the control action, which may result in jerky movements or instability.
We show how partial force control can eliminate these discontinuities.

The key idea is to control the forces associated to the constraints that are about to be added to/removed from the constraint set.
Namely, before lifting the right foot off the ground, we regulate its contact force to zero, while moving the COM over the left foot.
Similarly, when the right foot impacts the ground, we make its contact force slowly raise from zero to an appropriate value (i.e. the weight of the robot), while moving the COM over the right foot.
We report here the overall control hierarchy, in priority order: 
\begin{itemize}
\item constraints, either both feet (12 DoFs) or left foot (6 DoFs);
\item force control, right foot (6 DoFs); 
\item position control, COM ground projection (2 DoFs); 
\item position control, right foot (3 DoFs); 
\item position control, posture (29 DoFs).
\end{itemize}
Table~\ref{table:test2_timeline1} describes which tasks/constraints were active during the different phases of the test, and it briefly summarizes the task references.
\begin{table}[!t] 
\caption{Timeline of Test 2 using partial force control.}
\centering 
\begin{tabular}{p{2.1cm} | p{1.1cm} p{1.0cm} p{1.1cm} p{1.1cm} } 
\hline 
      	 Task                   		&    $0-2 s$	&   $2-4 s$ 	&    $4-6 s$	&    $6-8 s$\\  
	 [0.5ex] \hline \rowcolor[gray]{.9}
	 Constraints			&	Left foot	& Left foot		&  Left foot 	&   Both feet 	\\ 
	 Right foot wrench		&	Decrease	&      		&  Increase  	&     			\\ \rowcolor[gray]{.9}
	 COM			        &	Move left	& Stay still		&  Move right 	&   Move right 	\\ 
	 Right foot pose		&	         	& Move forward &             	&   			 \\ 
[0.5ex] \hline 
\end{tabular} 
\label{table:test2_timeline1} 
\end{table}
\begin{table}[!t] 
\caption{Timeline of Test 2 without using partial force control.}
\centering 
\begin{tabular}{p{2.0cm} | p{1.1cm} p{1.1cm} p{1.1cm} p{1.1cm} } 
\hline 
      	 Task                  		&    $0-2 s$	&   $2-4 s$ 	&    $4-5 s$	&    $5-8 s$\\  
	 [0.5ex] \hline \rowcolor[gray]{.9}
	 Constraints			&	Both feet	& Left foot		&  Left foot 	&   Both feet 	\\ 
	 COM			        &	Move left	& Stay still		&  Move right 	&   Move right 	\\ \rowcolor[gray]{.9}
	 Right foot pose		&	         	& Move forward &             	&   			 \\ 
[0.5ex] \hline 
\end{tabular} 
\label{table:test2_timeline2} 
\end{table}
For comparison, we performed the same test using the method proposed in \cite{Righetti2011}, which is similar from a computational standpoint, but it does not allow to control the contact forces.
Table~\ref{table:test2_timeline2} reports the timeline of this second test.
In this case we had to reintroduce the constraints on the right foot before 6 s (i.e. at 5 s) because at 5.5 s the robot could no longer balance due to the lack of force on the right foot.

Whenever there were more than six constraints (i.e. during double-support phase), we used the technique described in Section~\ref{sec:force_opt} to minimize the moments and the tangential forces at the feet.
In particular, we have set the weight matrix $W_f^{-1} \in \R{12}{12}$ to a diagonal matrix with entries $\mat{ w_l*f_{R}^{-1} & w_l*f_L^{-1}}$, where \mbox{$w_l = \mat{10 & 10 & 0.1 & 10^3 & 10^3 & 10^2}$}, and $f_R, f_L$ are the absolute values of the normal forces at the right and left foot, respectively.
In this way we penalized the tangential moments the most, followed by the normal moments and the tangential forces.
Moreover, we penalized more forces and moments at the foot on which the normal force was lower.
This was fundamental to maintain the Zero Moment Point (ZMP) \cite{Vukobratovic2004} inside the foot surface, especially when moving the COM away from the central position.

\begin{figure}[!tbp]
   \centering
   \includegraphics[width=0.5\textwidth]{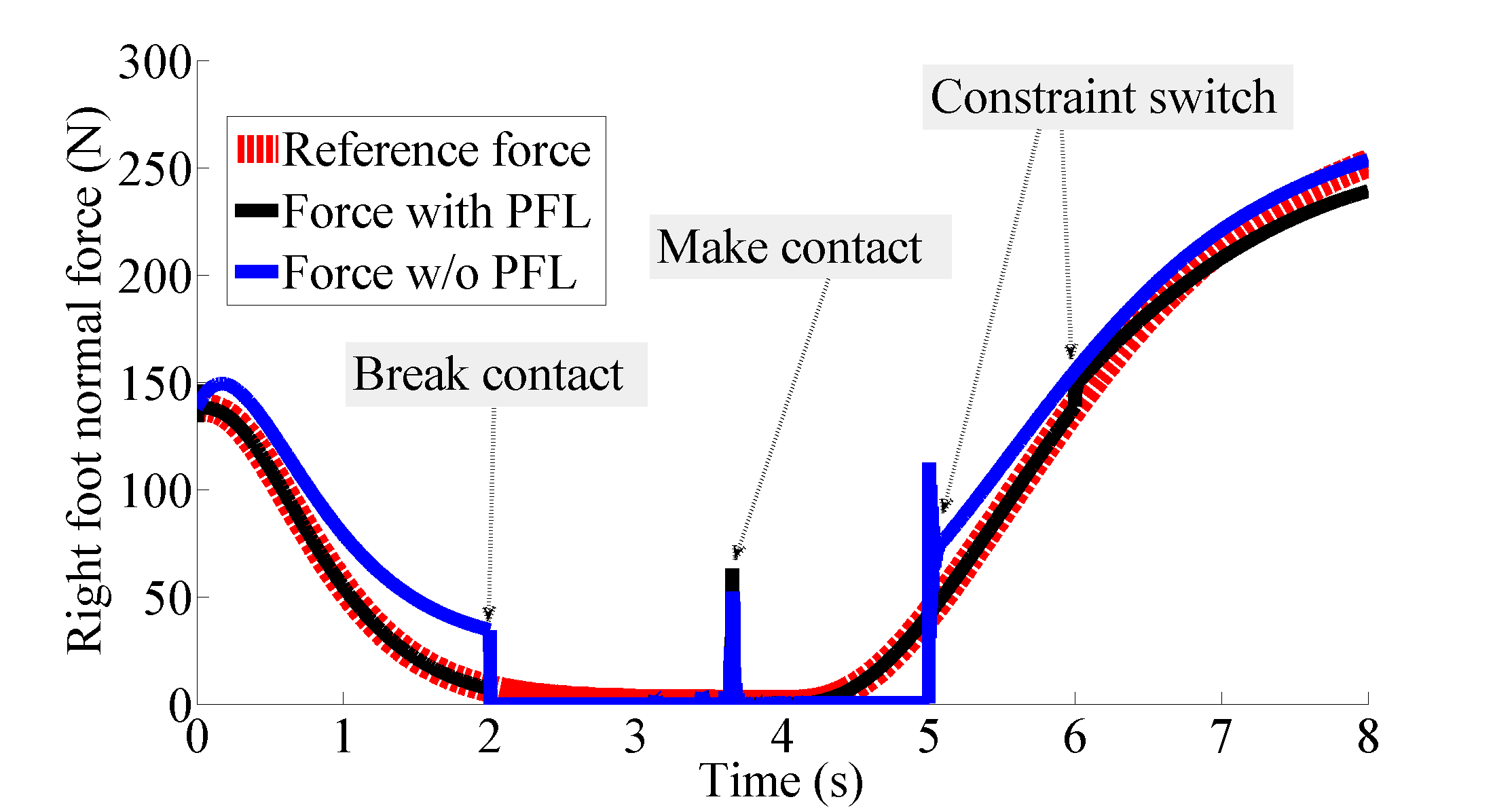}
   \caption{Test 2. Comparison of the normal contact force at the right foot when walking with/without using Partial Force Control. At 2 s the robot lifts the right foot off the ground. Right before 4 s the right foot makes contact with the ground. At 5 s and 6 s the number of constraints changes from 6 to 12.}
   \label{fig:fz_walk}
\end{figure}

Fig.~\ref{fig:fz_walk} shows the different normal contact forces at the right foot, obtained using the two approaches.
The force trajectory is almost continuous when using partial force control, whereas there are large discontinuities at 2 s and 5 s when we did not control the contact forces.
Moreover, thanks to partial force control, there is almost no discontinuity in the foot ZMPs when breaking the contact (i.e. 2 s) and at the switch of the number of constraints (i.e. 6 s).
On the contrary, when not controlling the foot force, there is a large discontinuity in the ZMP of the left foot at 2 s and 5 s.
The force discontinuity at the impact (right before 4 s) is independent of the control law: we commanded a desired foot position below the ground level, so the foot impacts the ground with nonzero velocity.
\section{Conclusions}
\label{sec:conclusions}
We proposed a reformulation of the constrained optimization arising in multi-task position/force control of constrained floating-base mechanical systems.
We derived a sparse analytical solution of the constraints of the problem, which exploits the structure of the equations of motion of the system.
The resulting unconstrained optimization has a reduced computational cost (about $20$ times faster for a humanoid) and completely decouples motion and force control. 
Moreover, the new formulation does not require calculating the mass matrix of the robot.
Other techniques based on inverse-dynamics projections \cite{Aghili2005} present similar computational complexity, but they do not allow for direct force control.

Our formulation is based on a physical insight in the dynamics of floating-base systems: if the constraint forces can accelerate the base in any direction, then the system can be seen as fully actuated.
We say that a robot that satisfies this condition is \emph{sufficiently constrained}.
In practice, this condition is often satisfied, as for the case of humanoids having at least one foot in flat contact with the ground.
To validate the theoretical results and demonstrate two possible applications we carried out simulations on a 23-DoF humanoid robot.

Future work consists of implementing the presented framework on a real humanoid robot.
Moreover, we are extending the framework to deal with inequalities and with the case of not \emph{sufficiently constrained} systems.

\section*{Acknowledgment}
The research in this paper was supported by the projects KoroiBot EU-FP7 and OSEO/Romeo2.



\bibliographystyle{IEEEtran}
\bibliography{IEEEabrv,references}

\end{document}